\documentclass[letterpaper]{article} 
\usepackage{aaai23}  
\usepackage{times}  
\usepackage{helvet}  
\usepackage{courier}  
\usepackage[hyphens]{url}  
\usepackage{graphicx} 
\usepackage{natbib}  
\usepackage{caption} 
\usepackage{algorithm}
\usepackage{algorithmic}
\usepackage{multirow}
\usepackage{amsfonts}
\usepackage{newfloat}
\usepackage{listings}
\DeclareCaptionStyle{ruled}{labelfont=normalfont,labelsep=colon,strut=off} 
\floatstyle{ruled}
\newfloat{listing}{tb}{lst}{}
\floatname{listing}{Listing}
\title{Few-Shot 3D Point Cloud Semantic Segmentation via Stratified Class-Specific Attention Based Transformer Network}
\author {
    Canyu Zhang,
    Zhenyao Wu, 
    Xinyi Wu, 
    Ziyu Zhao, 
    Song Wang
}
\affiliations {
     University of South Carolina, USA\\
    \{canyu, zhenyao, xinyiw, ziyuz\}@email.sc.edu, songwang@cec.sc.edu 
}

\usepackage{bibentry}

\begin{document}

\maketitle

\begin{abstract}
3D point cloud semantic segmentation aims to group all points into different semantic categories, which benefits important applications such as point cloud scene reconstruction and understanding. Existing supervised point cloud semantic segmentation methods usually require large-scale annotated point clouds for training and cannot handle new categories. While a few-shot learning method was proposed recently to address these two problems, it suffers from high computational complexity caused by graph construction and inability to learn fine-grained relationships among points due to the use of pooling operations. In this paper, we further address these problems by developing a new multi-layer transformer network for few-shot point cloud semantic segmentation. In the proposed network, the query point cloud features are aggregated based on the class-specific support features in different scales. Without using pooling operations, our method makes full use of all pixel-level features from the support samples. By better leveraging the support features for few-shot learning,  the proposed method achieves the new state-of-the-art performance, with 15\% less inference time, over existing few-shot 3D point cloud segmentation models on the S3DIS dataset and the ScanNet dataset. 
Our code is available at https://github.com/czzhang179/SCAT.
\end{abstract}

\begin{figure*}[ht]
  \includegraphics[width=\textwidth]{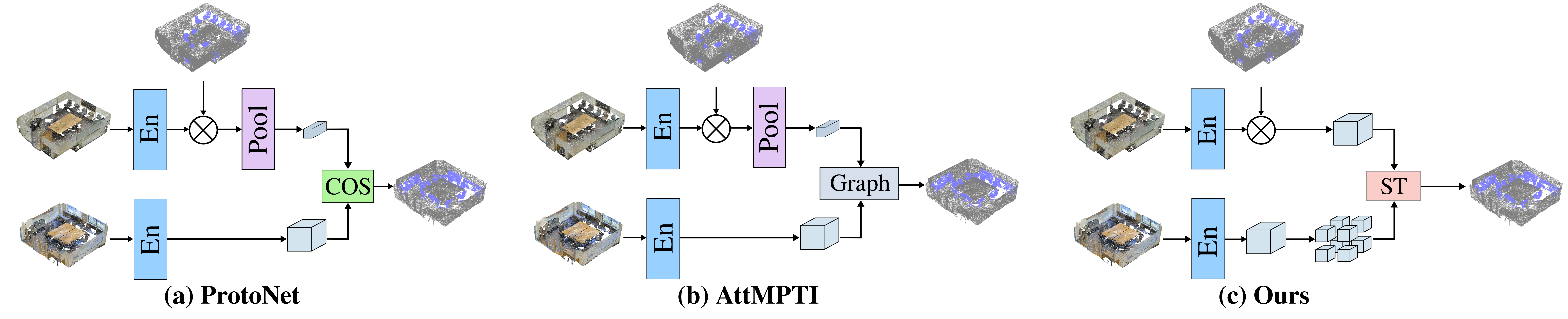} 

  \caption{Few-shot 3D point cloud semantic segmentation approaches.
  (a) ProtoNet ~\cite{snell2017prototypical} is proposed for 2D image few-shot semantic segmentation. Here we show its usage for 3D point clouds. The pooling operation is used to compute the prototype of the support samples for feature matching.
  (b) AttMPTI~\cite{zhao2021few} is the only existing few-shot based method for 3D point cloud semantic segmentation. It employs a graph network to measure the similarities among point features and prototypes.
  (c) Our proposed method leverages stratified transformer (ST) network to aggregate query features conditioned on the support features without using pooling operation.}
  \label{intro}
\end{figure*}

\section{Introduction}

Usually produced by a 3D scanner, (3D) point clouds play an important role in many computer vision and computer graphics applications. Given a complex scene, the captured point cloud may mix many different objects, structures and background. For better scene reconstruction and understanding, a critical step is \textit{point cloud semantic segmentation} which classifies each point into its underlying semantic category.
Recent state-of-the-art methods for point cloud semantic segmentation are mainly based on supervised deep learning~\cite{li2018pointcnn,qi2016pointnet,hu2019randla, liu2019densepoint, liu2021group,xu2021paconv}.
However, these methods have two major limitations: 1) they rely on large-scale training data, which require costly and laborious manual annotation, and 2) the trained model cannot segment new categories that are not seen during the training process.

By using a few labeled support examples as guidance, few-shot learning can well address these two limitations and has been shown to be effective in image segmentation~\cite{zhang2019canet,wang2019panet,vinyals2016matching}. Specifically, by taking the sample to be segmented as a query, one can compute a prototype using the labeled support data and then perform the segmentation of the query by measuring the distance between the prototype and query features similar. This can be extended to point-cloud segmentation as shown in Figure~\ref{intro}(a). 
Recently, AttMPTI~\cite{zhao2021few} made the first attempt to deploy few-shot method for point cloud semantic segmentation by leveraging a graph network to compute the distances between the prototype and point features in query, as shown in Figure~\ref{intro}(b).
However, this method is computationally expensive -- in practice, a point cloud usually contains a large number of points, leading to very large graph.
Another issue of this method comes from the use of a pooling operation for computing the prototype -- it limits the representation of the support samples~\cite{zhang2021few}. Since both global features and local features are essential for semantic segmentation, the prototype after pooling operation cannot fully capture diverse local semantic information from support and it may introduce noises when the feature is squeezed.

To address these issues, we propose a novel few-shot learning method for point cloud semantic segmentation by using a transformer network~\cite{dosovitskiy2020vit}, which has strong capability to learn long-range data dependencies via self-attention.
After extracting the features from both support and query point clouds, we apply transformer to explore the class-specific relationship between the support and query.
Without the pooling operation, we can get a dense relationship between all support category points and points in the query.
Recently, CNN-based models~\cite{huang2017wavelet,lotter2017multi,gong2014multi} show better performance via multi-scale technology. 
Following this idea, 
we represent the query (point cloud) in three different scales, and feed them into three transformer layers as shown in Figure~\ref{intro}(c). In this way, the distances between the query and support features are computed in a hierarchical way, which help the model learn both coarse- and fine-grained relationships.
Resulting features are then aggregated to produce the final perdition of the query sample.

For performance evaluation, we conduct comprehensive experiments on the widely used S3DIS~\cite{2017arXiv170201105A} and ScanNet~\cite{dai2017scannet} datasets in a variety of few-shot settings.
On the S3DIS dataset, our proposed network improves the mean-IoU by around 3\% in the one-shot setting and around 1\% in the five-shot setting. 
On the ScanNet dataset, our model can bring up the mean-IoU by around 2\% in the one-shot setting and around 1.5\% in the five-shot setting. Under the same setting, the proposed method can reduce the CPU inference time by around 15\%.

In summary, our contributions are listed as follows:
\begin{itemize}
    \item We introduce a new few-shot learning method for 3D point cloud semantic segmentation by introducing a stratified transformer network.
    \item We design a network to aggregate multi-scale features of query conditioned on the labeled support samples to better explore their relationships. 
    \item  We conduct comprehensive experiments to verify that our proposed method is more efficient and effective than existing few-shot learning methods and all the proposed network designs contribute to the performance improvement.
 
\end{itemize}

\section{Related Work}

\subsection{Point Cloud Semantic Segmentation}

A 3D point cloud is composed of a number of individual points with their $X$, $Y$, and $Z$ coordinates. Sometimes, each point also contains RGB color and intensity information depending on the sensors used to capture the point cloud.
Well-known public datasets of point clouds include ShapeNet~\cite{chang2015shapenet}, ModelNet40~\cite{wu20153d} and KITTI~\cite{Geiger2013IJRR}, with various manual annotations. 
Due to the rapid development of deep neural networks, many supervised point cloud processing methods have been proposed~\cite{atzmon2018point, hu2020jsenet, jiang2019hierarchical, landrieu2018large,lei2020seggcn, liu2019densepoint, liu2021group,xu2021paconv}. 

The goal of 3D point cloud semantic segmentation is to partition a point cloud scene into different meaningful semantic parts and it has been drawn much interest in the AI and vision communities. 
PointNet~\cite{qi2016pointnet} employs symmetric MLP layers to aggregate all point information for classification and segmentation. 
PointNet++~\cite{qi2017pointnet++} further learns local features with increasing contextual scales using metric space distance. 
PointCNN~\cite{li2018pointcnn} weights the input features associated with the points and permutes the points into a canonical order. 
DGCNN~\cite{wang2019dynamic} utilizes EdgeConv that is more aware of CNN-based high-level tasks to promote point cloud segmentation, and it can learn local and global point cloud information iteratively.
RandLA-Net~\cite{hu2019randla} uses a lighted-weight neural network to deal with large-scale point cloud data, with significant speed-up of the inference time.
All these methods require a number of point clouds with labeled ground truth for network training. Differently, our method follows the idea of few-shot learning which only uses a few labeled 3D point cloud samples as exemplars for training and testing. 

\subsection{Few-shot Semantic Segmentation}
Few-shot learning has been widely used for 2D image semantic segmentation in order to relieve the pixel-level annotation burden. OSLSM~\cite{shaban2017one} uses a fully convolutional network to segment images of new categories using few support data as guidance. 
PANet~\cite{wang2019panet} treats the few-shot segmentation as a metric learning problem and uses a prototype alignment network to explore information underlying support images. 
CANet~\cite{zhang2019canet} uses a dense comparison module to obtain a coarse prediction based on the support image, and several iterative optimization modules to refine this prediction.

Recently, the few-shot learning technique has also been used for 3D point cloud semantic segmentation -- AttMPTI ~\cite{zhao2021few} proposes a graph based few-shot network to segment the point cloud.
Inspired by AttMPTI, in this paper we further advance the few-shot point cloud semantic segmentation by developing a new method that aggregates point cloud features from different dimensions.

\begin{figure*}[th]
	\begin{center}
	    \hspace{-.5cm}\includegraphics[width=\linewidth]{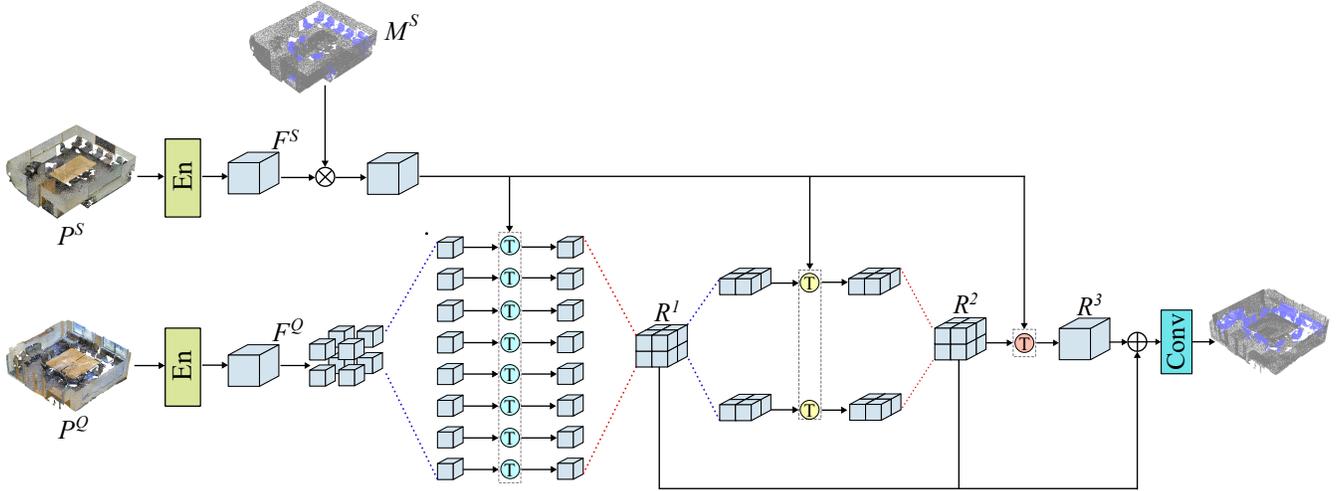}  \hspace{-.5cm}
		\caption{The framework of our proposed stratified class-specific attention based transformer network. It is composed of three layers of transformers which aggregate the query point cloud in different scales. } 
		\label{main}
	\end{center}
\end{figure*}

\section{Our Approach}

\subsection{Overview}
Given very few annotated point clouds (support), few-shot point cloud semantic segmentation aims to segment the corresponding semantic categories in query point cloud.
During training, our model tries to learn the best parameters for building the semantic relationship between the support and the query.
In testing process, the trained model can fit new categories and segment the query point cloud with the help of a few labeled support point clouds.

For convenience, we first introduce our proposed method in the one-shot (i.e., one support point cloud), one-way (i.e., one semantic class of interest and background) setting, and discuss its extension to more general multi-shot, multi-way setting later. As shown in Figure~\ref{main}, the inputs of the whole model consist of a query point cloud $P^{\mathcal Q}\in \mathbb{R}^{ N_q \times C}$ and a support point cloud $P^{\mathcal S}\in \mathbb{R}^{ N_s \times C}$  with corresponding label $M^{\mathcal S}\in \{0,1\}^{ N_s \times 1}$, where $N_q$ and $N_s$ are numbers of points in each point cloud and $C$ is the number of features for each point including the coordinates and possibly color channels. Query point cloud label $M^{\mathcal Q}\in \mathbb{R}^{ N_q \times 1}$ is the ground truth used for both training and evaluation. 
In the first step, two weight-sharing encoders are used to generate the support point cloud feature  $F^{\mathcal S}\in \mathbb{R}^{N_s \times D}$ and query point cloud feature $F^{\mathcal Q}\in \mathbb{R}^{N_q \times D}$ respectively, where $D$ is number of channels of the encoded feature. 

In our experiments, we choose DGCNN~\cite{wang2019dynamic} as our encoder, it consists of three EdgeConv layers to aggregate point features, which can learn the local and global 3D information iteratively.  
Then the newly proposed stratified transformer network builds the class-specific attention between the class-specific support feature $M^{\mathcal S}F^{\mathcal S}$ and $F^{\mathcal Q}$ at different scales. In the following sections, we first introduce the stratified architecture then elaborate on the class-specific attention based transformer. 
Finally, we extend the proposed method to the multi-shot and multi-way settings.

\subsection{Stratified Transformer Architecture} 

In few-shot point cloud segmentation, existing work ~\cite{zhao2021few} only builds single level correlation between the support branch and query branch, which is insufficient to construct accurate correlation. 
In our work, from a different perspective, we exploit both fine- and coarse- grained relationships between support and query by using a stratified transformer architecture, as shown in Figure \ref{main}.  
More specifically, the query point cloud is divided into several small point clouds according to the spatial location of the points, followed by three transformer layers to aggregate the point cloud features over multiple scales. 
In each layer, the transformer blocks are weight-sharing in order to reduce the number of parameters and boost the training speed.

To split the query point cloud, we first compute the XYX-axis aligned 3D rectangular bounding box that tightly covers all the query points. Then we uniformly divide this box into $2\times 2\times 2=8$ smaller boxes by splitting at midpoint along every side. 
The query points located in each smaller box form a new point cloud with corresponding feature $F^1_i\in \mathbb{R}^{N_i \times D}, i=1,2,...,8$, where $N_i$ is the number of points in the $i$-th small box. 
The transformers in the 1st layer aggregate each divided point-cloud features $F^1_i$ with class-specific support feature $M^{\mathcal S}F^{\mathcal S}$ to generate the class-specific attention, resulting in class-specific representation $R^1_i \in \mathbb{R}^{N_i \times D}$.
Then, all the class-specific representation $R^1_i$, are concatenated according to their original 3D locations to obtain output $R^1$. 
After that, the concatenated representation $R^1$ is equally divided into two parts along the z-axis (i.e., height), resulting in $F^2_i\in \mathbb{R}^{N_i \times D}, i=1,2$, corresponding to top- and bottom-half of the captured scene, respectively.
The top-bottom division is motivated by the indoor point cloud feature that the top and bottom parts usually contain the different objects. 
Each $F^2_i$ is aggregated with $M^{\mathcal S}F^{\mathcal S}$ via the second transformer layer and the outputs of top- and bottom-half of the scene are then concatenated to get the second class-specific representation $R^2$.

The last layer of transformer takes $R^2$ and $M^{\mathcal S}F^{\mathcal S}$ as inputs and generates the third query aggregated representation $R^3$.
With the above three transformer layers, $R^1$, $R^2$ and $R^3$ actually represent the features at multi-scales. We add them together and further apply two 1D convolutional layers and one $\mathtt{softmax}$ layer to get the final prediction $Prediction\in \mathbb{R}^{N \times 1}$. In training, we adopt the cross entropy $\mathtt{CE}$ loss:
\begin{equation}
    \mathcal Loss = \mathtt{CE}(Prediction, M^{\mathcal Q}).
    \label{1oss}
\end{equation}

\subsection{Class-specific Attention Based Transformer Block} \label{sec:att}

The core of few-shot semantic segmentation is to find the matching areas between query and support features. Previous works~\cite{zhao2021few} only utilize the single prototype after pooling operation, ignoring much information for matching. In this work, we use transformer blocks to build the point-wise relationship between the support and query point cloud features for better matching.

Firstly, each generated query point cloud feature $F  \in \{F^1_1, F^1_2, \cdots,  F^1_8, F^2_1, F^2_2, F_3\}$ will go through one self-attention layer as in the standard transformer \cite{vaswani2017attention}.
This self-attention layer can help our model focus on the feature distribution inside each generated $F$.
Then proposed specific-class attention based transformer takes the processed $F$ and class-specific support feature $M^{\mathcal S}F^{\mathcal S}$ as input.  
For class-specific attention based block, we construct $\mathbf Q$, $\mathbf K$ and $\mathbf V$ using three different linear layers:
\begin{equation}
    \mathbf Q =  FW_q, \qquad  \mathbf K = M^{\mathcal S}F^{\mathcal S}W_k, \qquad  \mathbf V = M^{\mathcal S}F^{\mathcal S}W_v,
\end{equation}
where $W_q, W_k, W_v \in \mathbb{R}^{D \times D_a}$ are learnable parameters which are used to project the input feature into the latent space, and $D_a$ is the number of channels in the transformer block. 

In each head, transformer multiplies $\mathbf Q$ and $\mathbf K$ to get the attention map,  which can represent the similarities between class-specific feature and the input query feature.  
Attention map is multiplied with $\mathbf V$ to transfer the attention map to original feature dimension.  
Mathematically, the $h$-th head can be calculated by:
\begin{equation}
    head_h = \mathtt{softmax} \left( \frac{\mathbf Q_h\mathbf K_h}{\sqrt{D_k}} \right) \mathbf V_h,
\end{equation}
where $D_k$ is the channel number of $\mathbf K_h$, and $\mathtt{softmax}$ means a row-wise softmax layer for attention normalization.
The output of all heads are combined together and added with the original input point cloud feature $F$.
The final class-specific representation $R$ is calculated by:
\begin{equation}
    R = F + \mathtt{Cat}(head_1, \cdots, head_H)W_o,
\end{equation}
where $H$ is the number of heads, and $W_o \in \mathbb{R}^{D_a \times D}$ is learnable, which projects the output to the original shape, and $\mathtt{Cat}$ is the concatenation operation.

\subsection{Multi-shot Multi-way Setting} \label{sec:shotway}

To extend the proposed method from one-shot to multi-shot setting, all we need to do is to update the class-specific support feature $M^{\mathcal S}F^{\mathcal S}$ to accommodate all the support point clouds. More specifically, we multiply the features of each support to its corresponding mask and then concatenate all of them, i.e., we get $M^{\mathcal S}F^{\mathcal S} \in \mathbb{R}^{W \times D}$, where $W$ is the total number of points of all the supports. 
Other than this, all the other steps, as well as the loss function, of the proposed method are exactly the same as the one-shot setting as discussed earlier.

For multi-way segmentation, i.e., segmenting for more than one semantic categories from a query point cloud, we consider each category independently.
For each category, we compute its class-specific feature and apply our method to get its segmentation prediction -- the transformers take different $\mathbf K$ and $\mathbf V$ for different category. 
The predictions of all the categories are then concatenated together and go through a $\mathtt{softmax}$ layer to get the final multi-way segmentation prediction.

\begin{table*}[th]

	\centering
	\renewcommand\arraystretch{1.2}
	\setlength{\tabcolsep}{1.2mm}{
		\begin{tabular}{l|c|c|c|c|c|c|c|c|c|c|c|c}
			\hline
		    \multirow{3}{*}{\textbf{Method}}
		    &\multicolumn{6}{c|}{\textbf{One-way}}&\multicolumn{6}{c}{\textbf{Two-way}}\\
		    \cline{2-13}
		    &\multicolumn{3}{c|}{\textbf{One-shot}}&\multicolumn{3}{c|}{\textbf{Five-shot}}&
		    \multicolumn{3}{c|}{\textbf{One-shot}}&\multicolumn{3}{c}{\textbf{Five-shot}}\\
		    \cline{2-13}
		    &$S^0$&$S^1$&$Mean$&$S^0$&$S^1$&$Mean$&$S^0$&$S^1$&$Mean$&$S^0$&$S^1$&$Mean$\\
		    \hline
		    ProtoNet&57.23&58.53 &57.88 &63.47&64.58&64.02&36.34&38.79&35.57&56.49&56.99&56.74\\ 
		 
		    \hline
		    MPTI&64.13&65.33&64.73&68.68 &68.04&68.45&52.27&51.58&51.88&58.93&60.65&59.75\\ 
		    \hline
		    AttMPTI &66.44&67.20&66.82&69.18&69.45&69.31 &53.77&55.94&54.96&61.67&67.02&64.35\\ 
		    \hline
		    3CAT-Ours&\underline{67.76}& \underline{68.89} & \underline{68.32}&\underline{69.43}&\underline{69.78}
		    &\underline{69.60}&\underline{53.96}&\underline{56.72}&\underline{55.34}&\underline{62.26}&\underline{68.94}&\underline{65.60}\\
		    
		    \hline
		    Ours &\textbf{69.37}& \textbf{70.56} &\textbf{69.96}  &\textbf{70.13}&\textbf{71.36}&\textbf{70. 74}&\textbf{54.92}&\textbf{56.74}&\textbf{55.83}&
		    \textbf{64.24}&\textbf{69.03}&\textbf{66.63}\\ 
		  
			\hline 
    	\end{tabular}}
    	
		\caption{Results of our network compared with ProtoNet, MPTI~\cite{zhao2021few},  AttMPTI~\cite{zhao2021few}, and our designed baseline -- 3CAT on S3DIS dataset. Bold represents the best results and underline means the second best results. $S_1$ indicates the swap of training and test sets in $S_0$.}
		\label{S3DIS}
\end{table*}

\begin{table*}[h]

	\centering
	\renewcommand\arraystretch{1.2}
	\setlength{\tabcolsep}{1.2mm}{
			\begin{tabular}{l|c|c|c|c|c|c|c|c|c|c|c|c}
			\hline
		    
		    \multirow{3}{*}{\textbf{Method}} &\multicolumn{6}{c|}{\textbf{One-way}}&\multicolumn{6}{c}{\textbf{Two-way}}\\
		    \cline{2-13}
		    &\multicolumn{3}{c|}{\textbf{One-shot}}&\multicolumn{3}{c|}{\textbf{Five-shot}}&
		    \multicolumn{3}{c|}{\textbf{One-shot}}&\multicolumn{3}{c}{\textbf{Five-shot}}\\
		    \cline{2-13}
		    &$S^0$&$S^1$&$Mean$&$S^0$&$S^1$&$Mean$&$S^0$&$S^1$&$Mean$&$S^0$&$S^1$&$Mean$\\
		    \hline
		    ProtoNet&50.21&51.45&50.83& 59.04&60.54& 59.79 &30.95& 33.92 &32.44 &42.01& 45.34& 43.68\\ 
		 
		    \hline
		    MPTI&52.13&57.59&54.86&62.13 &63.73&62.93&36.14& 39.27& 37.71& 43.59 &46.90 &45.25\\ 
		    \hline
		    AttMPTI &54.60&\underline{58.53}&56.56&62.86&64.91 &63.88&40.83& 42.55 &41.69 &50.32& 54.00& 52.16 \\ 
		    \hline
		   3CAT-Ours&\underline{55.41}& 58.52&\underline{56.96}& \underline{64.37}&\underline{65.06}&\underline{64.71}&\underline{43.86}&\underline{45.27}&\underline{44.56}&\underline{51.95}&\underline{56.63}&\underline{54.29}\\
		    
		    \hline
		    Ours&\textbf{56.49}& \textbf{59.22}&\textbf{57.85}&
		    \textbf{65.19}&\textbf{66.82}&\textbf{66.00}&\textbf{45.24}&
		    \textbf{45.90}&\textbf{45.57}&\textbf{55.38}&\textbf{57.11}&\textbf{56.24}\\ 
		  
			\hline 
    	\end{tabular}}
    	
		\caption{Results of our network compared with ProtoNet, MPTI~\cite{zhao2021few}  AttMPTI~\cite{zhao2021few} and our designed baseline --  3CAT on ScanNet dataset. Bold represents the best results and underline means the second best results. $S_1$ indicates the swap of training and test sets in $S_0$.}
		\label{scan}
		
\end{table*}

\section{Experiments}

\subsection{Datasets}

In this paper, we conduct the experiments on S3DIS and ScanNet datasets.
S3DIS~\cite{2017arXiv170201105A} dataset is composed of five large-scale indoor areas generated from three different buildings, which show diverse properties in architectural style and appearance.
It contains 12 semantic classes for semantic segmentation.
ScanNet~\cite{dai2017scannet} contains 1,513 point cloud scans from 707 special indoor scenes, which provides 20 semantic categories for segmentation.
Compared with S3DIS dataset, the ScanNet dataset provides more various room types, such as bathroom and living room.
And the point cloud scenes in ScanNet dataset are much more irregular, making the segmentation more difficult. 
For each dataset, we generate a training class set $C_{train}$ and a testing class set $C_{test}$, without overlap -- We perform cross-validation on each dataset using $C_{train}$ for training and validation, and $C_{test}$ for testing.

Since a point cloud of a scene usually contains large number of points, following previous work~\cite{qi2016pointnet, wang2019dynamic}, we divide the point cloud of a whole scene into many blocks using a non-overlapping sliding window in the $xy$ plane. In each block, $2,048$ points are randomly sampled, and we select our training set by only keeping the blocks containing more than $200$ points that belongs to the category of interest. 

\subsection{Implementation Details}
\paragraph{\textbf{Training}}

All models are implemented using PyTorch and trained on a Tesla v100 GPU.
Firstly, we pre-train the feature extractor (DGCNN) with additional MLP layers as classifier on the training dataset $C_{train}$ with point-wise supervision. 
For this pre-training stage, the batchsize is set to 32, and training epoch is 100. We use Adam optimizer ($ \beta_1$= 0.9, $ \beta_2$ = 0.999). The learning rate is set to 0.001.
In meta training process, we initialize the feature extractor by loading the pre-trained weight, and use the Adam optimizer ($ \beta_1$= 0.9, $\beta_2$ = 0.999) to update all parameters. Batchsize is set to 1. The initial learning rate is set to 0.001 and it is decayed by half every 5,000 iterations.  
In each iteration, training pair is constructed based on a randomly chosen category.

\paragraph{\textbf{Evaluation}}

In order to evaluate the performance of different models, we adopt mean Interaction over Union (mean-IoU) metric, which is widely used for evaluating point cloud semantic segmentation results. 
In the experiments, we evenly divide all categories into sets $C_{train}$ and $C_{test}$. In our tables, $S_0$ means using one set for training and the other for testing, while $S_1$ indicates the swapping of the training and testing sets.

\subsection{Baseline}

To better evaluate the performance of proposed stratified class-specific attention based transformer network, we choose three accumulated class-specific attention base transformers (3CAT) to process the support and query directly as our baseline method for comparison.

\subsection{Results}

In order to show the effectiveness of proposed network, we conduct one-shot and five-shot experiments in one-way and two-way settings respectively. We compare the results of our model with proposed baseline network 3CAT, previous few-shot point cloud semantic segmentation networks MPTI~\cite{zhao2021few} and AttMPTI~\cite{zhao2021few}, and the Prototypical Learning (ProtoNet) which uses cosine similarity to calculate the similarity between prototype and point features in query point cloud. 

The results of S3DIS dataset are shown in Table~\ref{S3DIS}. We can see that using our baseline 3CAT already gets a reasonably good result. 
In one-way one-shot setting, 3CAT bests ProtoNet by 10.44\%, and it outperforms MPTI by 3.59\% and AttMPTI by 1.50\%. In one-way five-shot setting, 3CAT also brings improvement compared with MPTI and AttMPTI.
In one-way setting, compared with 3CAT, the proposed stratified class-specific attention based transformer network improves the mean-IoU by 1.64\% in one-shot setting and 1.14\% in five-shot setting.
In two-way setting, 3CAT beats AttMPTI by 0.38\% in one-shot setting and 1.25\% in five-shot setting. Proposed method also earns the highest mean-IoU -- 0.49\% and 1.03\% over 3CAT, and 0.87\% and 2.28\% over AttMPTI, in one-shot and five-shot settings, respectively. 

\begin{figure}[th]
	\begin{center}
	    \hspace{-.5cm}\includegraphics[width=\linewidth]{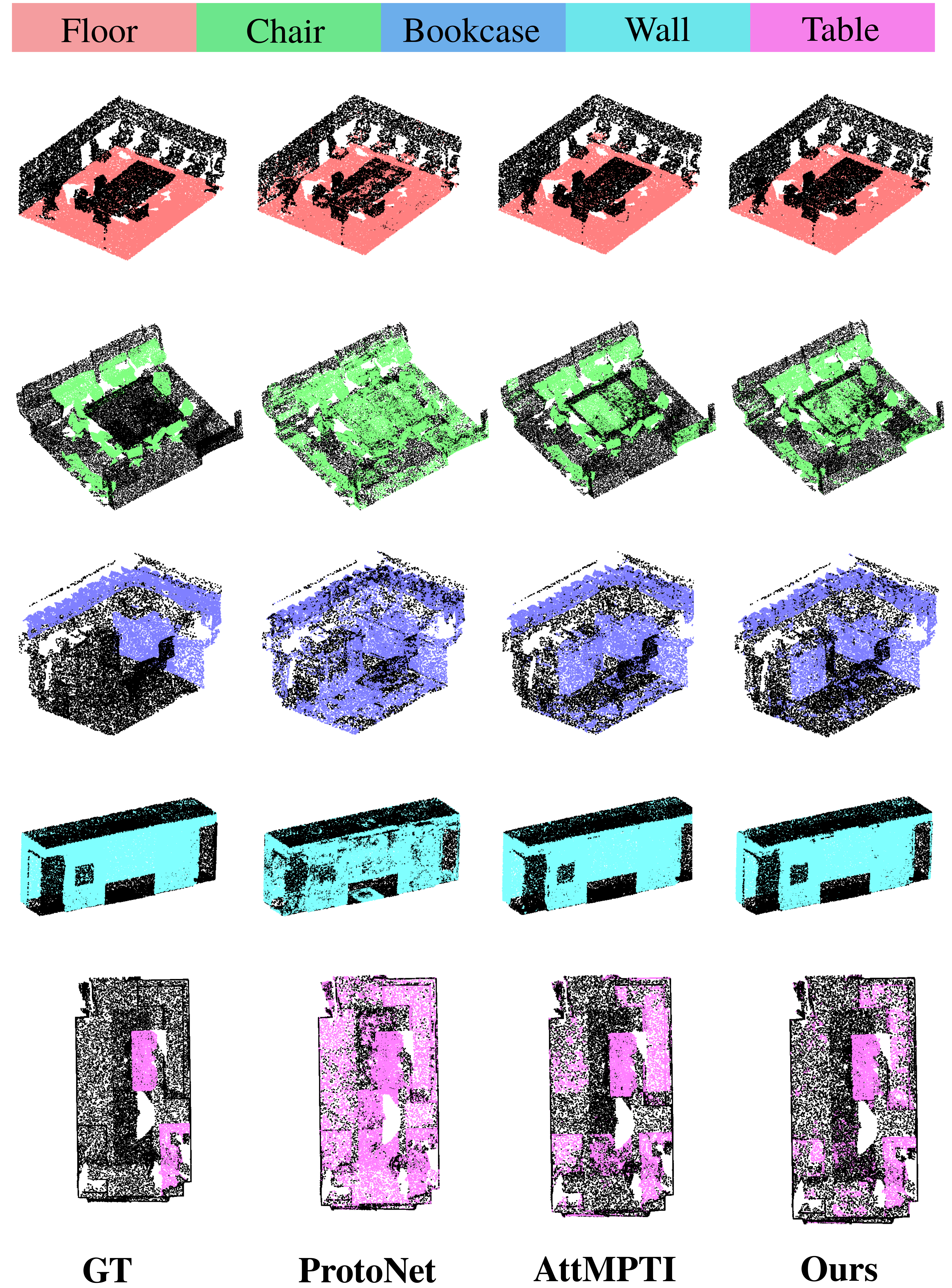} \hspace{-.5cm}
		\caption{Qualitative results of our method in one-way one-shot point cloud semantic segmentation on the S3DIS dataset, compared with the ground truth, ProtoNet and AttMPTI. From top to bottom, we show the performance on categories `floor' (red), `chair' (green), `bookshelf' (blue), `wall' (cyan-blue) and `table' (pink). }
		\label{s3disres}
	\end{center}
\end{figure}

The results of ScanNet dataset are shown in Table~\ref{scan}. Similarly, 3CAT can get reasonable mean-IoU.
Compared to ProtoNet, 3CAT gains around 6.13\% and 4.92\% improvements in one-way one-shot setting and one-way five-shot setting, respectively. 
Compared with AttMPTI, 3CAT earns around 0.40\% improvement in one-way one-shot setting and 0.83\% improvement in one-way five-shot setting.
Compared with 3CAT, our proposed method improves the mean-IoU by 0.89\% in one-way one-shot setting, and 1.29\% in one-way five-shot setting.
In two-way setting, 3CAT beats AttMPTI by 2.87\% and 2.13\% in one-shot setting and five-shot setting, respectively. 
Our proposed method still earns the highest mean-IoU -- it brings 1.01\% and 1.95\% improvements over 3CAT, and 3.88\% and 4.08\% improvements over AttMPTI in one-shot and five-shot settings, respectively. 
We can see that the proposed stratified structure can further improve the performance of 3CAT on few-shot point cloud semantic segmentation. Both experiments on S3DIS and ScanNet datasets show the effectiveness of proposed class-specific attention based transformer and our stratified architecture.

\begin{figure}[t]
	\begin{center}
	    \hspace{-.5cm}\includegraphics[width=\linewidth]{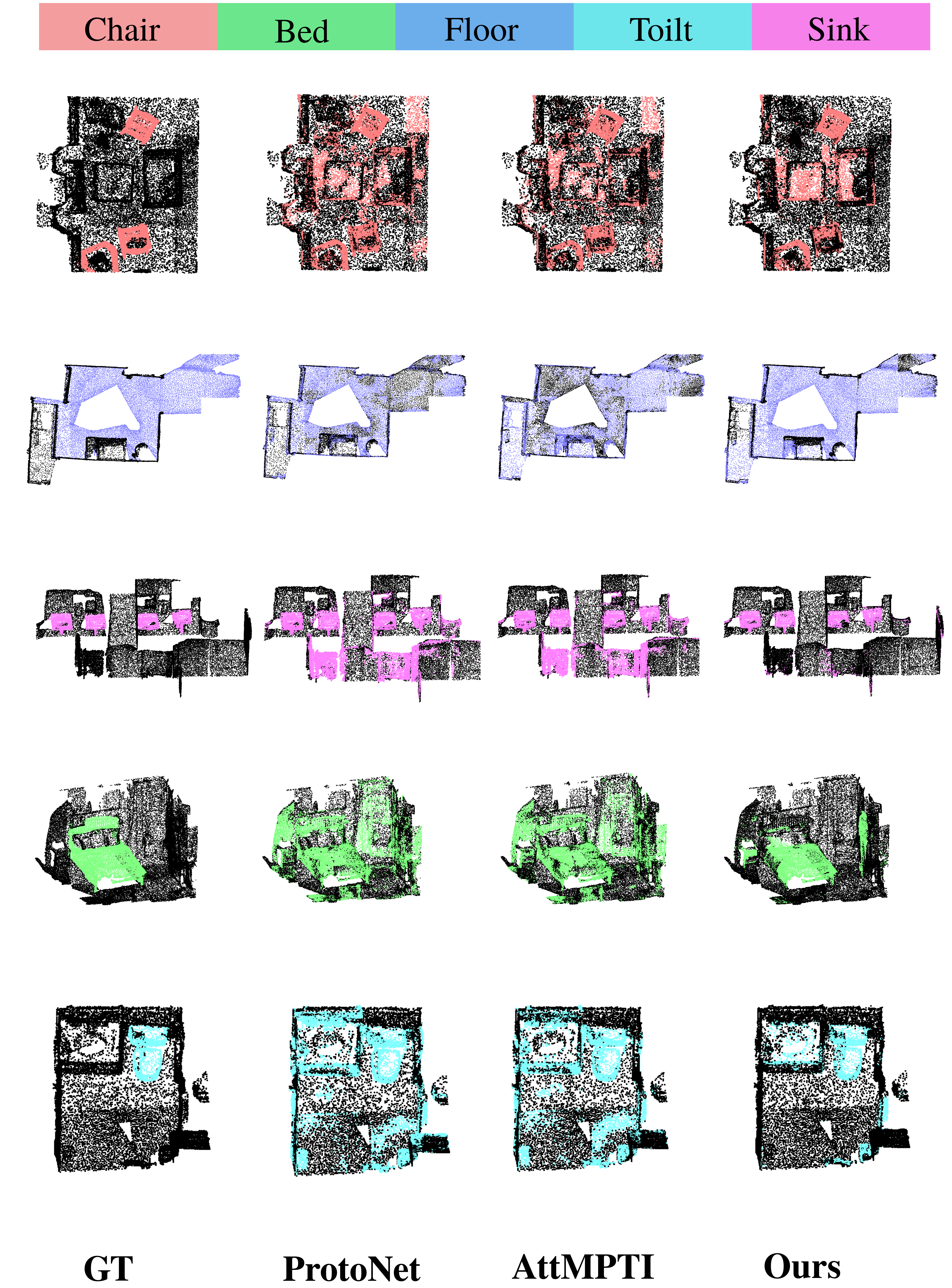} \hspace{-.5cm}
		\caption{Qualitative results of our method in one-way one-shot point cloud semantic segmentation on the ScanNet dataset. compared with the ground truth, ProtoNet and AttMPTI. From top to bottom, we show the performance on categories `chair' (red), `floor' (blue), `bathtub' (pink), `bed' (green) and `bed' (cyan-blue). }
		\label{scanres}
	\end{center}
\end{figure}

We further show qualitative results on S3DIS and ScanNet datasets in Figure~\ref{s3disres} and~\ref{scanres} respectively, under one-way one-shot setting, with the comparison with the qualitative results from AttMPTI, ProtoNet and ground truth (GT). 
As shown in Figure~\ref{s3disres}, ProtoNet and AttMPTI mis-classify some points of `table' category to the `floor' category (row 1) while our proposed method can avoid such mis-classification. 
When segmenting `bookcase' category, our model predicts the smallest number of wrong points. 
For the segmentation results on ScanNet dataset shown in Figure~\ref{scanres}, the predictions from ProtoNet and AttMPTI contain many irreverent points when segmenting `bathtub' (row 3) and `bed' (row 4) categories, while proposed method can bring a more precise category edge.
Thanks to well-designed class-specific attention based transformer and stratified structure, our model can segment each category more accurately.

\begin{figure}[h]
	\begin{center}
	    \hspace{-.5cm}\includegraphics[width=\linewidth]{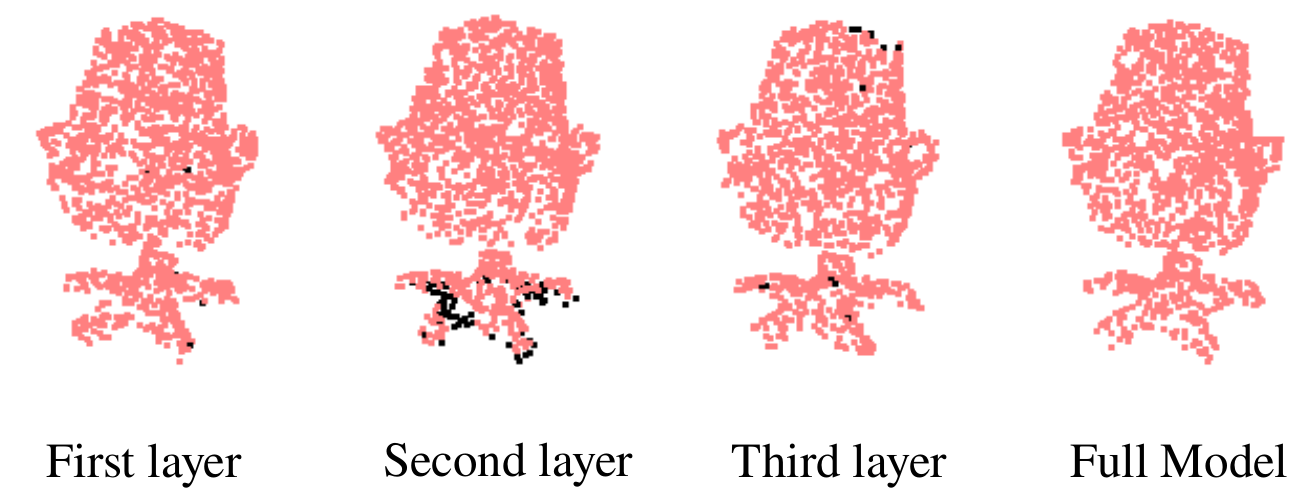} \hspace{-.5cm}
		\caption{Qualitative results on `chair' category from the proposed method and its variants. }
		\label{segf}
	\end{center}
\end{figure}

\subsection{Inference Time Analysis}

We compare the single point cloud inference time in one-way one-shot setting with ProtoNet, AttMPTI, MPTI, and 3CAT, and the corresponding results are shown in Table~\ref{time}.
For AttMPTI and MPTI, the size of constructed graph can be extremely large when processing large-scale point cloud, leading to more inference time. Compared with AttMPTI, 3CAT can save inference time by around 35\%. 
And compared with MPTI, 3CAT can save inference time by around 34\%.
Compared with AttMPTI and MPTI, our proposed model can reduce inference time by around 15\% and 14\%. 
It is still much more efficient than previous graph-based methods.

\begin{table}[h]

	\centering
	\renewcommand\arraystretch{1.05}
	\setlength{\tabcolsep}{1.2mm}{
			\begin{tabular}{l|c|c|c|c|c}
			\hline	
		    \textbf{Method} &ProtoNet& AttMPTI&MPTI&3CAT-Ours& Ours\\
		    \hline
		    \textbf{Time}(sec) &3,614& 4,762 &4,674&3,070&4,027 \\ 
		    \hline
    	\end{tabular}}
	\caption{Inference time (per point cloud, in average) of our proposed method compared with ProtoNet, AttMPTI, MPTI and 3CAT.}
	\label{time}
\end{table}

\subsection{Ablation Studies}

In this section, we report ablation studies to verify the effectiveness of the incorporated transformers, stratified architecture, 
class-specific representation, skip connection, and other experiment settings.
All experiments are conducted on S3DIS dataset in one-way one-shot setting unless claimed.

\paragraph{\textbf{Study on the Splitting Method }}
In order to justify the effectiveness of our stratified network, we compare the results of our models with 1) only splitting the query point cloud into $8$ small point clouds (\emph{First layer only});  2) only splitting the query point cloud into $2$ small point clouds (\emph{Second layer only}); 3) without any splitting (\emph{Third layer only}). 
We also test the performance of random splitting, which means we segment the whole query point cloud into small point clouds without considering their 3D locations. As shown in Table~\ref{seg}, the model considering all scales, i.e., with all three transformer layers, get the best performance, and it surpasses the models of \emph{First layer only} and \emph{Second layer only} by 0.71\% and 1.76\%, respectively.
Meanwhile using random splitting reduces the mean-IoU by 2.42\%. 

In Figure~\ref{segf}, we show the segmentation results on `chair' category using the proposed method and the above variants. The models only using the first, second or third transformer layer make wrong prediction on the points of chair back and legs, while the proposed method gets the best segmentation result.

\begin{table}[h]

	\centering
	\renewcommand\arraystretch{1.05}
	\setlength{\tabcolsep}{1.4mm}{
			\begin{tabular}{l|c|c|c}
			\hline
		    \textbf{Splitting choice}&$S^0$&$S^1$&$Mean$\\
		    \hline
			\hline
		    Third Layer Only &67.74& 68.88 & 68.31 \\
		    \hline
		    Second Layer Only &66.42& 69.98& 68.20\\
		    \hline
		    First Layer Only &68.34& 70.17& 69.25\\
		    \hline
		    Random Splitting &65.68& 69.41 &67.54 \\
		    \hline   
		    Our full model & \textbf{69.37}&  \textbf{70.56} & \textbf{69.96} \\ 
			\hline 
    	\end{tabular}}
        \caption{Results of the model using different splitting operations.}
        \label{seg}
\end{table}

\paragraph{\textbf{Study on the Class-specific Representation }}

In our method, we multiply the support point cloud feature $F^{\mathcal S}$ and corresponding mask $M^{\mathcal S}$ to get class-specific representation. To show the effectiveness of this design, we compare the performance with the the prototype-based approach. We get the prototype by multiplying support point cloud feature and mask, then calculating the average pooling result as $\mathbf K$ and $\mathbf V$ of all the transformers.
As shown in Table~\ref{abla}, using class-specific representation can increase mean-IoU by 1.82\% by keeping more category information compared with pooled prototype.

\begin{table}[t]

	\centering
	\renewcommand\arraystretch{1.05}
	\setlength{\tabcolsep}{1.4mm}{
			\begin{tabular}{l|c|c|c}
			\hline
		    \textbf{Method} &$S^0$&$S^1$&$Mean$\\
		    \hline
		    \hline
		    \textbf{Ours} &\textbf{69.37}& \textbf{70.56} &\textbf{69.96} \\ 
			\hline 
		    Using Prototype &67.58& 68.70& 68.14\\
		    \hline
		    Without Skip Connection &66.84& 68.33& 67.58\\
		    \hline
		    Using Single Head Attention &67.91& 69.22 &68.56 \\
		    \hline
    	\end{tabular}}
    	\caption{Results of ablation studies on class-specific attention, skip connection in transformer, multi-head attention.}
    	\label{abla}
\end{table}

\paragraph{\textbf{Study on the Skip Connection}}

In order to show the effectiveness of skip connection in our class-specific attention based transformer, we conduct ablation study by removing skip connection in all transformers. As shown in Table~\ref{abla}, using skip connection in transformer can improve the mean-IoU by 2.38\%. It shows that using skip connection can increase the performance of transformer network by reducing feature degradation.

\paragraph{\textbf{Study on the Multi-head Attention}}

In our experiments, we use a four-head attention in our transformer to capture rich information. We conduct ablation study to show the influence of different number of heads. The results are shown in Table~\ref{abla}. Compared with the transformer network using one-head attention, the transformer network using four-head attention can increase the mean-IoU by 1.40\%. This experiment shows the effectiveness of muti-head attention in our proposed method.

\section{Conclusion}

In this paper, we proposed a stratified class-specific attention based transformer network for few-shot 3D point cloud semantic segmentation. Compared with previous single prototype-based approach, the proposed class-specific transformer can keep more support category information to process the query point cloud features. Moreover, the stratified architecture is introduced to exploit both fine- and coarse-grained relationships between the support and query. Thorough experiments and ablation studies on popular S3DIS dataset and ScanNet dataset show the effectiveness of our proposed method and the design of each step. It outperforms MPTI, AttMPTI and the baseline 3CAT, with about 15\% less inference time than MPTI and AttMPTI. 

\section{Acknowledgements}

This work used GPUs provided by the NSF MRI-2018966.

\bibliography{aaai23}

\end{document}